\def\year{2021}\relax
\documentclass[letterpaper]{article} 
\usepackage{aaai21}  
\usepackage{times}  
\usepackage{helvet} 
\usepackage{courier}  
\usepackage[hyphens]{url}  
\usepackage{graphicx} 
\usepackage{graphicx}  
\usepackage{natbib}  
\usepackage{caption} 
\usepackage{latexsym}
\usepackage{amsmath}
\usepackage{amsfonts}
\usepackage{amssymb}
\usepackage{multirow}
\usepackage{adjustbox}
\usepackage[english]{babel} 
\usepackage{microtype}
\usepackage{arydshln}
\usepackage{color}

\title{Co-GAT: A Co-Interactive Graph Attention Network for Joint Dialog Act Recognition and Sentiment Classification}

\author{
	Libo Qin,
	Zhouyang Li,
	Wanxiang Che,\thanks{corresponding author.}
	Minheng Ni,
	Ting Liu\\
}
\affiliations{
	Research Center for Social Computing and Information Retrieval, \\
	Harbin Institute of Technology, Harbin, China \\
	
	\{lbqin, zhouyangli, car, mhni, tliu\}@ir.hit.edu.cn
}
\begin{document}

\maketitle

\begin{abstract}
In a dialog system, dialog act recognition and sentiment classification are two correlative tasks to capture speakers’ intentions, where dialog act and sentiment can indicate the explicit and the implicit intentions separately.
The dialog context information (\textit{contextual information}) and the \textit{mutual interaction information} are two key factors that contribute to the two related tasks.
Unfortunately,
none of the existing approaches consider the two important sources of information simultaneously.
In this paper, we propose a \textbf{C}\textbf{o}-Interactive \textbf{G}raph \textbf{A}ttention Network (Co-GAT) to jointly perform the two tasks.
The core module is a proposed co-interactive graph interaction layer where a \textit{cross-utterances connection} and a \textit{cross-tasks connection} are constructed and iteratively updated with each other, achieving to consider the two types of information simultaneously.
Experimental results on two public datasets show that our model successfully captures the two sources of information and achieve the state-of-the-art performance.
 In addition, we find that the
contributions from the contextual and mutual interaction information do not
fully overlap with contextualized word representations (BERT, Roberta, XLNet).

\end{abstract}

\section{Introduction}
\label{Introduction}
Dialog act recognition (DAR) and sentiment classification (SC) are two correlative tasks to correctly understand speakers’ utterances in a dialog system \cite{cerisara-etal-2018-multi,lin2020discovering,qin2020dcr_net}.
DAR aims to attach semantic labels to each utterance in a dialog which represent the underlying intentions.
Meanwhile,
SC can detect the sentiments in utterances which can help to capture speakers’ implicit intentions.
More specifically, DAR can be treated as a sequence classification task that maps the utterances sequence $(u_{1},u_{2}...,u_{N})$ to the corresponding utterances sequence DA label ($y_{1}^{d},y_{2}^{d},...,y_{N}^{d}$), where $N$ is the number of utterances in dialog.
Similarly, SC can be also seen as an utterance-level sequence classification problem to predict the corresponding utterances sentiment label ($y_{1}^{s},y_{2}^{s},...,y_{N}^{s}$).
\begin{figure}[t]
	\centering
	\includegraphics[scale=0.4]{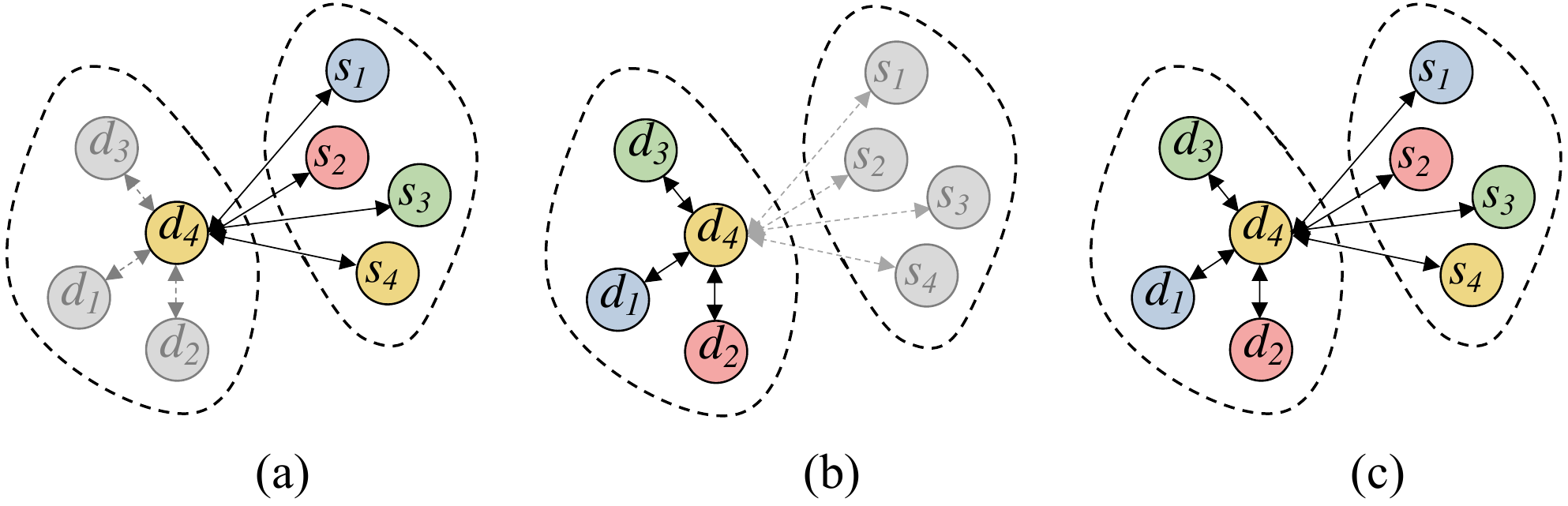}
	\caption{\label{constract} 
		Methods for Joint DAR and SC. Previous work either incorporate \textit{the mutual information} (a), or leverage \textit{the contextual information} (b).
		Our co-interactive graph interaction method can leverage both the two sources of information (c).
		$s$ denotes sentiment representation and $d$ denotes dialog act representation.
	}
	\label{example}
\end{figure}

\begin{figure*}[t]
	\centering
	\includegraphics[scale=0.66]{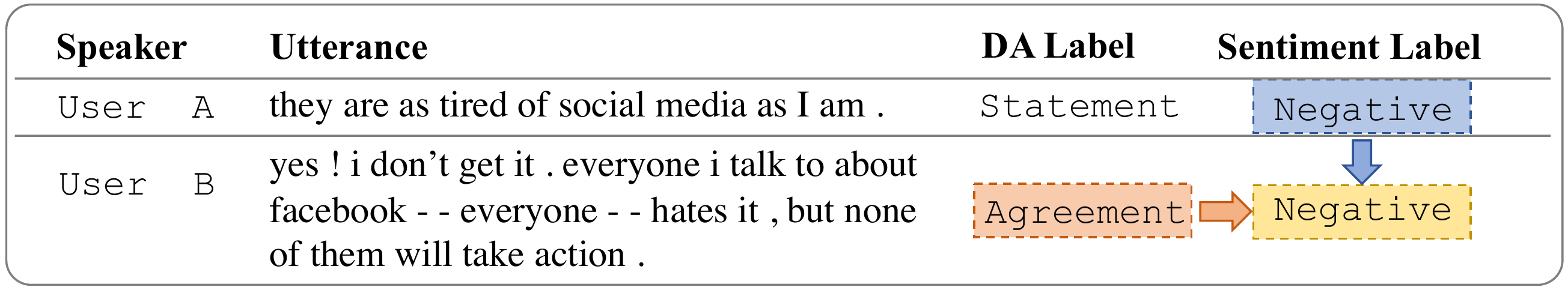}
	\caption{\label{corpus-examples}  A snippet of a dialog sample from the Mastodon corpus and each utterance has a corresponding DA label and a sentiment label. (DA represents Dialog Act). 
		The blue color segment represents \textit{the contextual information} and the red segment denotes \textit{the mutual interaction information} while the yellow segment represents the target label to predict. 
	}
	\label{example1}
\end{figure*}
Intuitively, there are two key factors that contribute to the dialog act recognition and sentiment prediction. One is \textit{the mutual interaction information} across two tasks and the other is \textit{the contextual information} across utterances in a dialogue.
As illustrated in Figure~\ref{corpus-examples},  to predict the sentiment label of \texttt{User B}, annotated with \texttt{Negative},  its mutual interaction information (i.e. \texttt{Agreement} DA label) and contextual information (i.e. sentiment label of \texttt{User A}) contribute a lot to the final prediction.
The reason is \texttt{Agreement} means 
the \texttt{User B} agrees with previous \texttt{User A} and hence the \texttt{User B} sentiment label tends to be \texttt{Negative}, the same with the \texttt{User A} sentiment label (\texttt{Negative}). 
Similarly, knowing the mutual sentiment interaction information and the contextual information also contributes to the DA prediction.
Thus, it's critical to take the two sources of information into account.

To this end,  \citet{cerisara-etal-2018-multi} proposes a multi-task framework to jointly model the two tasks, which can implicitly extract the shared mutual interaction information, but fail to effectively capture the contextual information, which is shown in Figure~\ref{constract}(a).
\citet{kim2018integrated} explicitly leverage the previous act information to guide the current DA prediction, which  captures the contextual information, which is shown in Figure~\ref{constract}(b).
However, the model ignores the mutual interaction information, which can be used for promoting the two tasks.
Recently, \citet{qin2020dcr_net} propose a \textit{pipeline} method (\textit{DCR-Net}) to incorporate the two types of information. 
In \textit{DCR-Net}, a hierarchical encoder is proposed to capture the contextual information, followed then by a relation layer to consider the mutual interaction information.
Although DCR-Net has achieved good performance, we argue that the \textit{pipeline} method suffers from one major issue:  \textit{two information are modeled separately}.
This means 
the updated process of two types of information are totally isolated, resulting in one type of information can not propagate  another type of information in the updated process, which is not effective enough for leveraging knowledge across 
utterances and tasks.
In general, the existing models either consider only one source of information, or employ the above two types of information with \textit{pipeline} modeling method. This leaves us with a question: \textit{Can we simultaneously model the mutual interaction and contextual information in a unified framework to fully 
	incorporate them ?}

Motivated by this, we propose 
a \textbf{Co}-Interactive \textbf{G}raph \textbf{A}ttention Network (Co-GAT) for joint dialog act recognition and sentiment classification.
The core module is a proposed \textit{Co-Interactive} graph interaction layer, which achieves to fully use the aforementioned two sources of information simultaneously.
In \textit{Co-Interactive} graph, we perform a dual-connection interaction where a \textit{cross-utterances connection}  and a \textit{cross-tasks connection} are constructed and iteratively updated with each other, which is shown in Figure~\ref{example}(c). 
More specifically, the \textit{cross-utterances connection}, where each utterance connects other utterances in the same dialog, is used for capturing the contextual information.
 The \textit{cross-tasks connection}, where node in one task connects all nodes in another task, is used for making an explicit interaction to obtain the mutual interaction information.
Further, the \textit{cross-utterance connection}  and \textit{cross-task connection} are updated simultaneously and integrated into a unified graph architecture, achieving to answer the proposed question: \textit{each utterance node can be updated simultaneously with the contextual information and mutual interaction information}.

We conduct experiments on two real-world benchmarks including
Mastodon dataset \cite{cerisara-etal-2018-multi} and Dailydialog dataset \cite{li-etal-2017-dailydialog}. Experimental results show that our model achieves significant and consistent improvements as compared to all baseline models by successfully aggregating the mutual interaction information and contextual information.
On Mastodon dataset, our model gains 3.0\% and 1.9\% improvement on F1 score on SC and DAR task, respectively. On Dailydialog dataset, we also obtain 5.6\% and 0.3\% improvement.
In addition, we explore the 
pre-trained model (BERT, Roberta, XLNet) \cite{devlin-etal-2019-bert,liu2019roberta,yang2019xlnet} in our framework.

In summary, the main contributions of our work are concluded as follows:
\begin{itemize}
	\item We make the first attempt to simultaneously incorporate \textit{contextual information} and \textit{mutual interaction information} for joint dialog act recognition and sentiment classification.
	
	\item We propose a co-interactive graph attention network where a \textit{cross-tasks connection} and \textit{cross-utterances connection} are constructed and iteratively updated with each other, achieving to model simultaneously incorporate contextual information and mutual interaction information.
	
	\item  Experiments on two publicly available datasets show that our model obtains substantial improvement and achieves the state-of-the-art performance. In addition, our framework is also beneficial when combined with pre-trained models (BERT, Roberta, XLNet). 
\end{itemize}

To make our experiments reproducible, we will make our code and data publicly available at 
\url{https://github.com/RaleLee/Co-GAT}.

\section{Approach}
\label{Approach}
\begin{figure*}[t]
	\centering
	\includegraphics[scale=0.85]{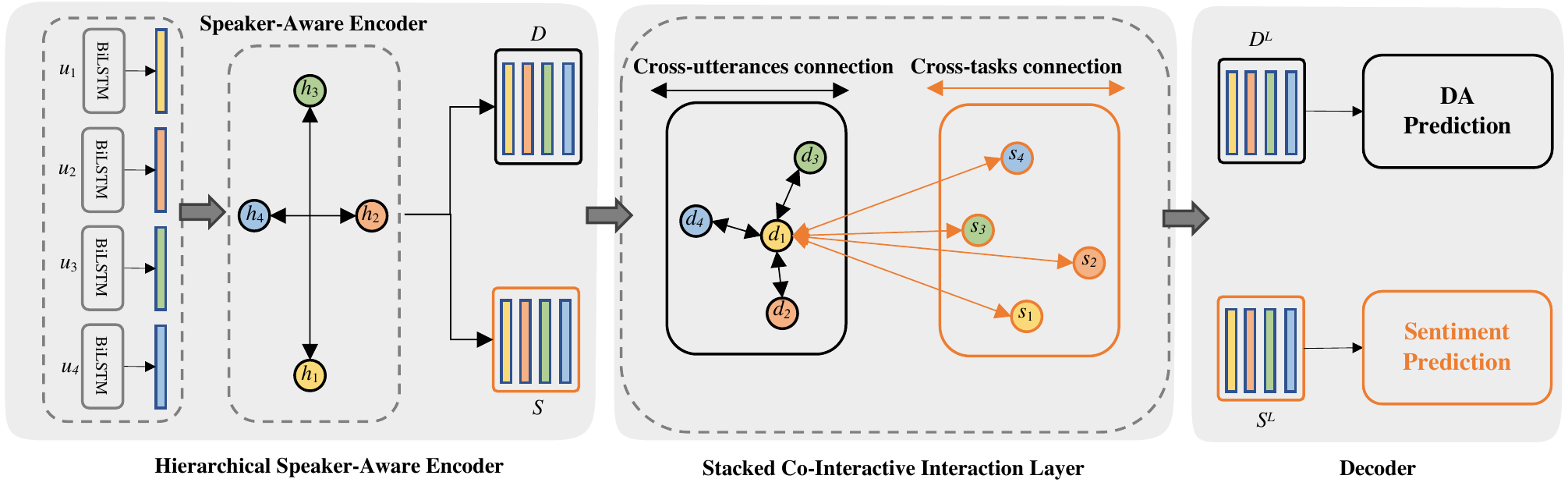}
	\caption{\label{model:propagate}  
		The illustration of our proposed framework, which consists of a hierarchical speaker-aware encoder, a stacked graph-based interaction layer and two separate decoders.
	}
	\label{fig:framework}
\end{figure*}

In this section, we describe the architecture of our framework, as illustrated in Figure~\ref{fig:framework}.
It is mainly composed of three components: a shared hierarchical speaker-aware encoder, a stack of co-interactive graph layer to simultaneously incorporate the contextual information and mutual interaction information, and two separate decoders for dialog act and sentiment prediction. In the following paragraph, we first describe the vanilla graph attention network and then the details of other components of framework are given.

\paragraph{Vanilla Graph Attention Network}
A graph attention network (GAT) \cite{velivckovic2017graph} is a variant of graph neural network \cite{scarselli2009graph}.
It propagates features from other neighbourhood's information to the current node and has the advantage of automatically determining the importance and relevance between the current node with its neighbourhood.

In particular, for a given graph with $N$ nodes, one-layer GAT take the initial node features $\tilde{\boldsymbol{H}} = \{\tilde{\boldsymbol{h}}_{1},\ldots, \tilde{\boldsymbol{h}}_{N}\}, \tilde{\boldsymbol{h}}_{n} \in \mathbb{R}^{F}$ as input, aiming at producing more abstract representation, $\tilde{\boldsymbol{H}}^\prime = \{\tilde{\boldsymbol{h}}^{\prime}_{1},\ldots,\tilde{\boldsymbol{h}}^{\prime}_{N}\}, \tilde{\boldsymbol{h}}^{\prime}_{n} \in \mathbb{R}^{F^\prime}$, as its output. The graph attention operated on the node representation can be written as:
\begin{eqnarray}
	\tilde{\boldsymbol{h}}^{\prime}_i =  \sigma\big(\sum_{j \in \mathcal{N}_i} \alpha_{ij} \boldsymbol{W}_h\tilde{\boldsymbol{h}}_{j}\big) \,,
\end{eqnarray}
where $\mathcal{N}_i$ is the first-order neighbors of node $i$ (including $i$) in the graph; ${F}$ and ${F^\prime}$ are the input and output dimension; $\boldsymbol{W}_h \in \mathbb{R}^{F^\prime \times F}$ is the trainable weight matrix and  $\sigma$ represents the nonlinearity activation function. 

The weight $\alpha_{ij}$ in above equation is calculated via an attention process, which models the importance of each $h_j$ to $h_i$:
\begin{eqnarray}
	\alpha_{ij} = \frac{\exp(\mathcal{F}(\tilde{\boldsymbol{h}}_{i}, \tilde{\boldsymbol{h}}_{j}))}{\sum_{j^\prime \in \mathcal{N}_i} \exp{(\mathcal{F}(\tilde{\boldsymbol{h}}_{i}, \tilde{\boldsymbol{h}}_{j^\prime}))}} \,, 
\end{eqnarray}
where $\mathcal{F}$ is an attention function.

In our experiments, following \citet{velivckovic2017graph}, the attention function can be formulated as:
\begin{eqnarray}
	\mathcal{F}(\tilde{\boldsymbol{h}}_{i}, \tilde{\boldsymbol{h}}_{j}) &= \operatorname{LeakyReLU}\left(\mathbf{a}^\top[\boldsymbol{W}_h\tilde{\boldsymbol{h}}_i\|\boldsymbol{W}_h\tilde{\boldsymbol{h}}_j]\right) \,,
\end{eqnarray}
where  $\mathbf{a} \in \mathbb{R}^{2F^\prime}$ is the trainable weight matrix.

In addition, to stabilize the learning process of self-attention, GAT extend the above mechanism to employ \textit{multi-head attention} from~\citep{NIPS2017_7181}: 
\begin{equation} \label{eq:multi-head}
	\tilde{\boldsymbol{h}}^{\prime}_i =  \mathop{||}_{k=1}^{K} \sigma\big(\sum_{j \in \mathcal{N}_i} \alpha_{ij}^k \boldsymbol{W}_h^k \tilde{\boldsymbol{h}}_{j}\big) \,,
\end{equation}
where $K$ is the number of heads, $\alpha_{ij}^k$ is the normalized attention weight at $k$ head and $\mathop{||}$ is concatenation operation and $K$ is the number of heads. 

Finally, following \citet{velivckovic2017graph}, we employ averaging instead of concatenation to get the final prediction results.

\subsection{Hierarchical Speaker-Aware Encoder}\label{sec:encoder}

In our framework, a hierarchical speaker-aware encoder is shared across the dialog act recognition and sentiment classification to leverage the implicit shared knowledge.
Specially, it consists of a bidirectional LSTM (BiLSTM) \cite{hochreiter1997long}, which captures temporal relationships within the words, followed by a speaker-aware graph attention network \cite{velivckovic2017graph} to incorporate the speaker information.
\subsubsection{Utterance Encoder with BiLSTM}
Given a dialog $C$ = $(u_{1},...,u_{N})$ consists of a sequence of $N$ utterances and the $t$-th utterance $u_{t}$ = $(w^{t}_{ 1},..., w^{t}_{n})$ which consists of a sequence of $n$ words,
the encoder first maps the tokens in $w^{t}_{i}$ to vectors
with embedding function $\phi^{\text{emb}}$.
Then, BiLSTM reads it 
forwardly from $w^{t}_{ 1}$ to $w^{t}_{n}$ and backwardly from $w^{t}_{ n}$ to $w^{t}_{ 1}$ to produce a series of context-sensitive hidden states $\boldsymbol{H} = \{\boldsymbol{h}^{t}_{1}, \boldsymbol{h}^{t}_2, \ldots, \boldsymbol{h}^{t}_{n}\}$.
Equations are as follows: 
\begin{align} \label{BiLSTM}
	& \overrightarrow{\boldsymbol{h}^{t}_{i}} = 	\overrightarrow{\operatorname{LSTM}}(\phi^{\text{emb}} (w^{t}_{i}), 	\overrightarrow{\boldsymbol{h}^{t}_{i-1}}), t \in [1, n] \,, \\
	&\overleftarrow{\boldsymbol{h}^{t}_{i}} = 	\overleftarrow{\operatorname{LSTM}}(\phi^{\text{emb}} (w^{t}_{i}), 	\overleftarrow{\boldsymbol{h}^{t}_{i+1}}), t \in [n, 1] \,, \\
	&\boldsymbol{h}^{t}_{i} = [\overrightarrow{\boldsymbol{h}^{t}_i},\overleftarrow{\boldsymbol{h}^{t}_i}].&
\end{align}

Then, the last hidden state $\boldsymbol{h}^{t}_{n}$ can be seen as the utterance $u_{t}$ representation $\boldsymbol{e}_{t}$ (i.e., $\boldsymbol{e}_{t}$ = $\boldsymbol{h}^{t}_{n}$).
Hence, the sequentially encoded feature of $N$ utterances in $C$ can be represented as $\boldsymbol{E}$ = ( $ \boldsymbol{ e }_1$, \dots, $ \boldsymbol{ e }_N$). 
\subsubsection{Speaker-Level Encoder}
We propose to use a speaker-aware graph attention network to leverage the speaker information, which enables the model to better understand how the emotion and act intention change within the same speaker \cite{ghosal-etal-2019-dialoguegcn}.
We build graphical structures over the input utterance sequences to explicitly incorporate the speaker information into the graph attention network, and construct the graph in the following way,

\textbf{Vertices:} Each utterance in the conversation is represented as a vertex.
Each vertex is initialized with the corresponding sequentially encoded feature vector $\boldsymbol{ e }_i$, for all $i \in [1,2,...,N]$. 
We denote this vector as the vertex feature. Hence, the first layer states vector for all nodes is  $\boldsymbol{E}$ = ( $ \boldsymbol{ e }_1$, \dots, $ \boldsymbol{ e }_N$). 

\textbf{Edges:} Since we aim to model the speaker information in a dialog explicitly, vertex $i$ and vertex $j$ should be connected if they belong to the same speaker.
More specifically, $\boldsymbol{ A}$ is an adjacent matrix $\boldsymbol{ A} \in \mathbb{R}^{N\times N}$ with $\boldsymbol{ A}_{ij}=1$ if they're from the same speaker and $\boldsymbol{ A}_{ij}=0$ otherwise\footnote{In our paper, we only consider the first-order neighbors to alleviate the overfitting problem.},
By doing this, the speaker features can be propagated from neighbour nodes to the current node.
In particular, the aggregation process can be rewritten as:
\begin{equation} \label{eq:multi-head}
	\tilde{\boldsymbol{e}}^{\prime}_i =  \mathop{||}_{k=1}^{K} \sigma\big(\sum_{j \in \mathcal{S}_i} \alpha_{ij}^k \boldsymbol{W}_h^k \tilde{\boldsymbol{e}}_{j}\big) \,,
\end{equation}
where $\mathcal{S}_i$ represents the nodes that belong to the same speaker with $i$ node.

After stacking $m$ layer, we obtain the speaker-aware encoding features ${\boldsymbol{ E}^{m}}$ = $(\boldsymbol{ e}^{m}_{1},\dots,\boldsymbol{ e}^{m}_{N})$.
Following \citet{qin2020dcr_net}, we first apply separate BiLSTM over act information and sentiment information separately to make them more task-specific, which can be written as $\boldsymbol{ D}^{0}$ = BiLSTM (${\boldsymbol{ E}^{m}}$) and $\boldsymbol{ S}^{0}$ = BiLSTM (${\boldsymbol{ E}^{m}}$).
$\boldsymbol{ D}^{0}$ = $(\boldsymbol{ d}^{0}_{1},\dots,\boldsymbol{ d}^{0}_{N})$ and $\boldsymbol{ S}^{0}$ = $(\boldsymbol{ s}^{0}_{1},\dots,\boldsymbol{ s}^{0}_{N})$ can be seen as the initial shared representations of dialog act and sentiment.

\subsection{Stacked Co-Interactive Graph  Layer}\label{sec:interaction}
One core advantage of our framework is modeling the contextual information and mutual interaction information into a unified graph interaction architecture and updating them simultaneously.
Specially, we adopt a graph attention network (GAT) to model the interaction process with the \textit{cross-tasks connection} and \textit{cross-utterances connection}.
Graph interaction structure has been shown effective on various of NLP tasks \cite{lu-li-2020-gcan,chai-wan-2020-learning,qin-etal-2020-agif,qin2020multi}.
We construct the graph in the following way,

\textbf{Vertices:} Since we model the interaction between the two tasks in graph architecture, we have 2$N$ nodes in the graph where $N$ nodes for sentiment classification task and the other $N$ nodes for dialog act recognition task.
We use the speaker-aware encoding representation $\boldsymbol{ D}^{0}$ and $\boldsymbol{ S}^{0}$ to initialize our corresponding sentiment and dialog act node vertices, respectively. Thus, we obtain the initialization node representation $\boldsymbol{ H}^{0}$ = [$\boldsymbol{ D}^{0}$;$\boldsymbol{ S}^{0}$] = $[\boldsymbol{ d}_{1}^{0},\dots,\boldsymbol{ d}_{N}^{0},\boldsymbol{ s}_{1}^{0},\dots,\boldsymbol{ s}_{N}^{0}]\in$ $\mathbb{ R}^{2N \times d}$, where $d$ represents the dimension of vertice representation.

\textbf{Edges:} In the graph, there exist two types of edges. 

\textbf{\textit{cross-utterances connection}}: 
We construct the \textit{cross-utterances connection} where node $i$ should connect to its context utterance node to take the contextual information into account.
More specifically, we denote the graph adjacent matrix as $\boldsymbol{ A} \in$ $\mathcal{ R}^{2N\times2N}$, where the $\boldsymbol{ A}^{I}_{i,j}$ = 1 if they are in the same dialogue.

\textbf{\textit{cross-tasks connection}}: 
The \textit{cross-tasks connection} is constructed where node $i$ connects to all another task node to explicitly leverage the mutual interaction information, where $\boldsymbol{ A}^{I}_{i,j}$ = 1 when they belongs to the different tasks.

By doing this, we model the two sources of information in a unified graph interaction framework with \textit{cross-utterances connection} and \textit{cross-tasks connection}.
In particular, we use $\boldsymbol{ d}_{i}^{(l)}$ and $\boldsymbol{ s}_{z}^{(l)}$ to represent dialog act representation of 
node $i$ and sentiment representation of node $z$ in the $l$-th layer of the graph, respectively. 
For $\boldsymbol{ d}_{i}^{(l)}$, the graph interaction update process can be formulated as:
\begin{eqnarray} \label{eq:multi-head2}
{\boldsymbol{d}}^{(l+1)}_i =  \mathop{||}_{k=1}^{K} \sigma\big(\sum_{j \in \mathcal{D}_i} \alpha_{ij}^k \boldsymbol{W}_h^k {\boldsymbol{d}}_{j}^{(l)} +\sum_{j \in \mathcal{A}_i} \alpha_{ij}^k \boldsymbol{W}_h^k {\boldsymbol{s}}_{j}^{(l)} \big)\,, 
\end{eqnarray}
where $\sum_{j \in \mathcal{D}_i} \alpha_{ij}^k \boldsymbol{W}_h^k {\boldsymbol{d}}_{j}^{(l)}$ is the \textit{cross-utterances connection} to integrate the contextual information while $\sum_{j \in \mathcal{A}_i} \alpha_{ij}^k \boldsymbol{W}_h^k {\boldsymbol{s}}_{j}^{(l)}$ denotes the \textit{cross-tasks connection} for incorporating the mutual interaction information.

Similarly,  the graph interaction update process for $\boldsymbol{s}_{i}^{(l)}$ can be formulated as:
\begin{eqnarray} \label{eq:multi-head3}
	{\boldsymbol{s}}^{(l+1)}_i = \mathop{||}_{k=1}^{K}  \sigma\big( \sum_{j \in \mathcal{A}^{\prime}_i}  \alpha_{ij}^k \boldsymbol{W}_h^k {\boldsymbol{s}}_{j}^{(l)} + \sum_{j \in \mathcal{D}^{\prime}_i}\alpha_{ij}^k \boldsymbol{W}_h^k {\boldsymbol{d}}_{j}^{(l)} \big)\,.
\end{eqnarray}

\subsection{Decoder for Dialog Act Recognition and Sentiment Classification}\label{sec:decoder}
In order to learn deep features, we apply a stacked graph attention network with multiple layers. After stacking $L$ layer, we obtain a final updated feature representation $\boldsymbol{ E}^{L}$ = $[\boldsymbol{ D}^{L};\boldsymbol{ S}^{L}]$ including: ${{\boldsymbol{ D}}}^{L}$ = (${\boldsymbol{ d}}_{1}^{L}$, ...,${\boldsymbol{ d}}_{N}^{L}$) and  ${{\boldsymbol{ S}}}^{L}$ = (${\boldsymbol{ s}}_{1}^{L}$, ...,${\boldsymbol{ s}}_{N}^{L}$). 
Then, we perform linear transform and LSTM upon the ${{\boldsymbol{ S}}}^{L}$ and ${{\boldsymbol{ D}}}^{L}$ to make the representation more task-specific, where the ${{\boldsymbol{ S}}}^{L'}$ = {Linear} $({{\boldsymbol{ S}}}^{L})$ and ${{\boldsymbol{D}}}^{L'}$ = {LSTM} {$({{\boldsymbol{ D}}}^{L}$)}.
We then adopt separate decoder to perform dialog act and sentiment prediction, which can be denoted as follows: 
\def\softmax{\mathop{\rm softmax}}
\begin{eqnarray}
	{\boldsymbol{ y}}_{t}^{d} &=& \softmax ({\boldsymbol{ W}}^{d}{ \boldsymbol{ d} }_{t}^{L'} + {\boldsymbol{ b}}_{d}), \\
	{\boldsymbol{ y}}_{t}^{s} &=& \softmax ({\boldsymbol{ W}}^{s}{\boldsymbol{s} }_{t}^{L'} + {\boldsymbol{ b}}_{s}),
\end{eqnarray}
where ${\boldsymbol{ y}}_{t}^{d} $ and ${\boldsymbol{ y}}_{t}^{s}$ are the predicted distribution for dialog act and sentiment respectively; ${\boldsymbol{ W}}^{d}$ and ${\boldsymbol{ W}}^{s}$ are transformation matrices; ${\boldsymbol{ b}}_{d}$ and ${\boldsymbol{ b}}_{s}$ are bias vectors.
\subsection{Joint Training}
The dialog act recognition and sentiment classification objective are formulated as:

\begin{eqnarray}
	\mathcal{L}_{1} \triangleq - \sum_{i=1}^{N}\sum_{j=1}^{N_S}{{\hat{{y}}_{i}^{(j,s)}}\log \left({{y}}_{i}^{(j,s)}\right)},\\
	\mathcal{L}_{2} \triangleq - \sum_{i=1}^{N}\sum_{j=1}^{N_D}{{\hat{{y}}_{i}^{(j,d)}}\log \left({{y}}_{i}^{(j,d)}\right)},
\end{eqnarray}
where ${\hat { {{y}} } _ { i } ^ { d } }$ and 
$ {\hat { {{y}}} _ { i } ^ { s } }$ are gold act label and 
gold sentiment label separately; $N_D$ is the number of dialog act labels; $N_S$ is the number of sentiment labels and $N$ is the number of utterances.

Following \citet{qin-etal-2019-stack}, the dialog act recognition and sentiment classification can be considered jointly, the final joint objective is:
\begin{equation}
	\mathcal { L } _ { \theta } =  \mathcal{ L }_{1} + \mathcal{ L }_{2}.
\end{equation}

\section{Experiments}
\label{Experiments}
\begin{table*}[t]
	\small
	\centering
	\begin{adjustbox}{width=1.0\textwidth}
		\begin{tabular}{l|ccc|ccc|ccc|ccc} 
			\hline 
			\multirow{3}*{\textbf{Model}} & \multicolumn{6}{c}{\textbf{Mastodon}} & \multicolumn{6}{c}{\textbf{Dailydialog}} \\ 
			\cline{2-13} 
			~ & \multicolumn{3}{c|}{SC} & \multicolumn{3}{c|}{DAR} & \multicolumn{3}{c|}{SC} & \multicolumn{3}{c}{DAR} \\ 
			\cline{2-13}  
			~ & F1 (\%) & R (\%) & P (\%) & F1 (\%) & R (\%) & P (\%) & F1 (\%) & R (\%) & P (\%) & F1 (\%) & R (\%) & P (\%) \\  
			\hline

			HEC \cite{kumar2018dialogue} & - &-   & -&56.1  &55.7  &56.5 &-  &- &- &77.8  &76.5  &77.8  \\  
			
			CRF-ASN \cite{chen2018dialogue} & - & -  &- &55.1  &53.9  &56.5 &-  &- & -&76.0  &75.6  &78.2 \\ 
			
			CASA \cite{raheja-tetreault-2019-dialogue} & - &-   &- &56.4  &57.1  &55.7 & - &- &- &78.0  &76.5  &77.9 \\ 
			\hline 
			
			DialogueRNN \cite{majumder2019dialoguernn} &41.5 &42.8 & 40.5 & - & - & - & 40.3 & 37.7 & 44.5 & - & - & - \\ 
			DialogueGCN \cite{ghosal-etal-2019-dialoguegcn}
			&42.4 &43.4 & 41.4 & - & - & - & 43.1 & 44.5 & 41.8 & - & - & - \\ 
			\hline 
			JointDAS \cite{cerisara-etal-2018-multi} & 37.6 & 41.6 & 36.1 & 53.2 & 51.9 & 55.6 & 31.2 & 28.8 & 35.4 & 75.1 & 74.5 & 76.2 \\  
			
			IIIM \cite{kim2018integrated} & 39.4 & 40.1 & 38.7 & 54.3 & 52.2 & 56.3 & 33.0 & 28.5 & 38.9 & 75.7 & 74.9 & 76.5 \\ 
			
			DCR-Net + Co-Attention \cite{qin2020dcr_net} & {45.1} & {47.3} & {43.2} & {58.6} & {56.9} & {60.3} & {45.4} & {40.1} & 56.0 & {79.1} & {79.0} & {79.1} \\  
			\hline 
			Our model & \bf{48.1*} & \bf{53.2*} & \bf{44.0*} & \bf{60.5*} & \bf{60.6*} & \bf{60.4} & \bf{51.0*} & \bf{45.3*} & \bf{65.9*} & \bf{79.4} & {78.1} & \bf{81.0*} \\   
			\hline  
			
		\end{tabular}
	\end{adjustbox}
	\caption{Comparison of our model with baselines on Mastodon and Dailydialog datasets. SC represents Sentiment Classification and DAR represents Dialog Act Recognition. The numbers with * indicate that the improvement of our model over all baselines is statistically significant with $p<0.05$ under t-test.}
	\label{table:over_all}
	
\end{table*}

\subsection{Datasets}
We conduct experiments on the benchmark Dailydialog \cite{li-etal-2017-dailydialog} and Mastodon \cite{cerisara-etal-2018-multi}.
On Dailydialogues dataset, we follow the same format and partition as in \citet{li-etal-2017-dailydialog}.
The dataset contains 11,118 dialogues for training, 1,000 for validation and 1,000 for testing.
On Mastodon dataset, it includes 239 dialogues for training, 266 dialogues for testing.
We keep the train/test partition unchanged.\footnote{The two datasets are available in \url{http://yanran.li/dailydialog} and \url{https://github.com/cerisara/DialogSentimentMastodon
}}
\subsection{Experimental Settings}
In our experiment setting, dimensionality of all hidden units are 256. And the dimensionality of the embedding is 800 and 128 for Mastodon and Dailydialog, respectively.
L2 regularization used on our model is $1\times 10^{-8}$.
In addition, we add a residual connection in graph attention network layer for reducing overfitting.
We use Adam \cite{kingma-ba:2014:ICLR} to 
optimize the parameters in our model and adopt
the suggested hyper-parameters for optimization.
We set the stacked number of GAT as 2 on Mastdon dataset and 3 on Dailydialog dataset.
For all experiments, we pick the model which works best on the dev set, and then evaluate it on the test set. 
All experiments are conducted at GeForce RTX 2080Ti. The epoch number is 300 and 100 for Mastodon and Dailydialog, respectively.

\subsection{Baselines}
We compare our model with several of state-of-the-art baselines including: 1) the separate dialog act recognition models:  HEC, CRF-ASN and CASA. 2) the separate sentiment classification models: DialogueGCN and DialogueRNN.
3) the joint models including: JointDAS, IIIM and DCR-Net.
We briefly describe these  baseline models below: 1) \textbf{{HEC}} \cite{kumar2018dialogue}: This work uses a hierarchical Bi-LSTM-CRF model for dialog act recognition, which capture both kinds of dependencies including word-level and utterance-level. 2) \textbf{{CRF-ASN}} \cite{chen2018dialogue}: This model proposes a crf-attentive structured network for dialog act recognition, which dynamically separates the utterances into cliques. 3) \textbf{{CASA}} \cite{raheja-tetreault-2019-dialogue}: This work leverages a context-aware self-attention mechanism coupled with a hierarchical deep neural network. 
4) \textbf{{DialogueRNN}} \cite{majumder2019dialoguernn}:
This model proposes a RNN-based neural architecture for emotion detection in a conversation to keep track of the individual party states throughout the conversation and uses this information. 
5) \textbf{DialogGCN}
\cite{ghosal-etal-2019-dialoguegcn}:
This model proposes a dialogue graph convolutional
network to leverage self
and inter-speaker dependency of the interlocutors to model conversational context for emotion recognition
6) \textbf{{JointDAS}} \cite{cerisara-etal-2018-multi}: This model uses a multi-task modeling framework for joint dialog act recognition and sentiment classification. 7) \textbf{{IIIM}} \cite{kim2018integrated}: This work proposes an integrated neural network model which simultaneously identifies speech acts, predicators, and sentiments of dialogue utterances.
8) \textbf{DCR-Net}: This model proposes a relation layer to explicitly model the interaction between the two tasks and achieves the state-of-the-art performance.

\begin{table}[t]
	
	\centering
	\begin{adjustbox}{width=0.48\textwidth}
		\begin{tabular}{l|c|c|c|c} 
			\hline 
			\multirow{2}*{\textbf{Model}} & \multicolumn{2}{c}{\textbf{Mastodon}} & \multicolumn{2}{c}{\textbf{Dailydialog}} \\ 
			\cline{2-5} 
			~ & \multicolumn{1}{c|}{SC (F1)} & \multicolumn{1}{c|}{DAR (F1)} & \multicolumn{1}{c|}{SC (F1)} & \multicolumn{1}{c}{DAR (F1)} \\ 
			\hline    
			
			without cross-tasks connection & 46.1 & 58.1 & 49.7 & 78.2\\
			without cross-utterances connection & 44.9 & 58.7 & 48.1 & 78.2\\
			separate modeling & 46.7 & 58.4 & 45.6 & 78.3 \\ 
			co-attention mechanism & 46.5 & 59.4 & 46.2 & 79.1 \\ 
			without speaker information & 46.4 & 59.0  & 47.6 & 79.2 \\
			\hline 
			Our model &\textbf{48.1} & \textbf{60.5}& \textbf{51.0} & \textbf{79.4} \\
			\hline
			
		\end{tabular}
	\end{adjustbox}
	\caption{Ablation study on Mastodon and Dailydialog test datasets.}
	\label{table:no_relation}
\end{table} 

\begin{figure*}[t]
	\centering
	\begin{minipage}{0.33\linewidth}\centering
		\includegraphics[width=4.8cm]{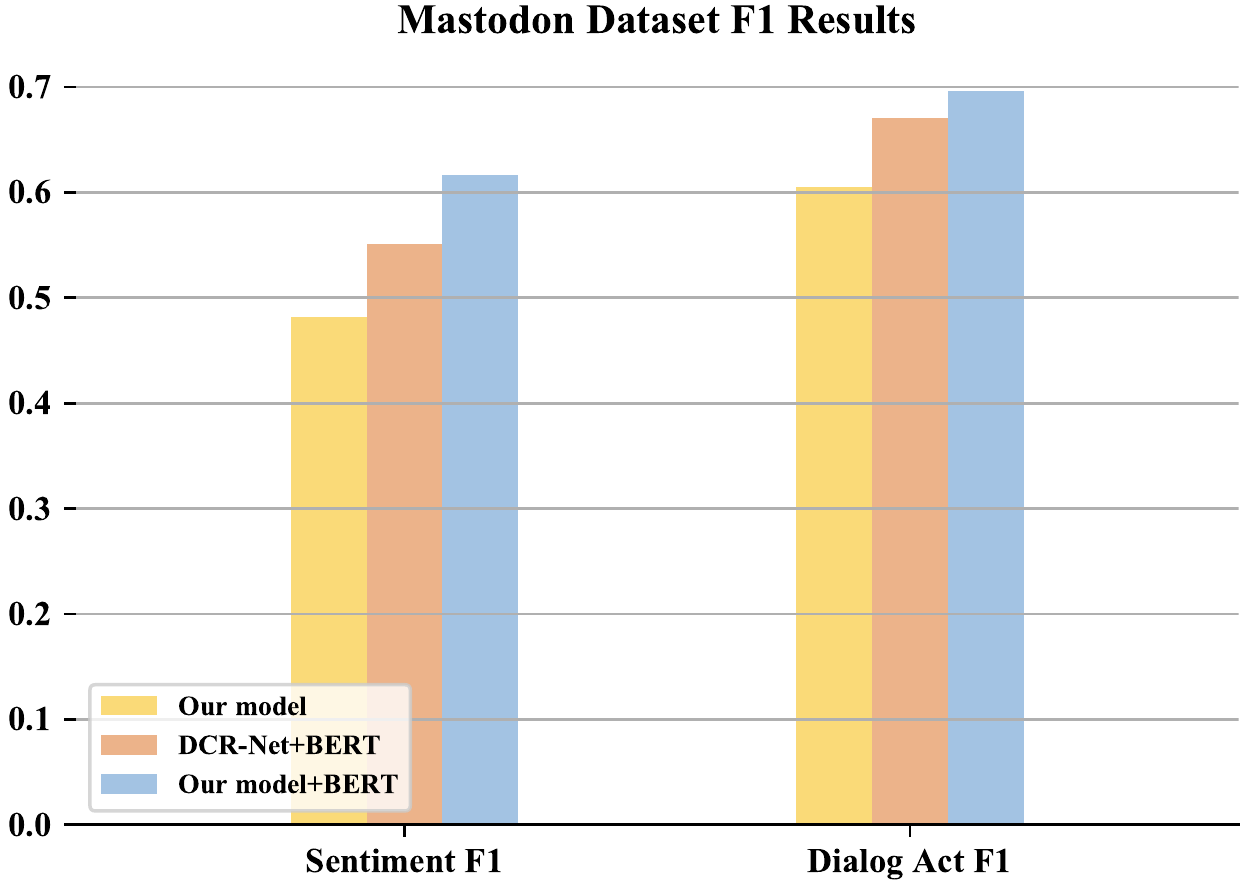}\\
		(a) 
	\end{minipage}
	\begin{minipage}{0.3\linewidth}\centering
		\includegraphics[width=4.8cm]{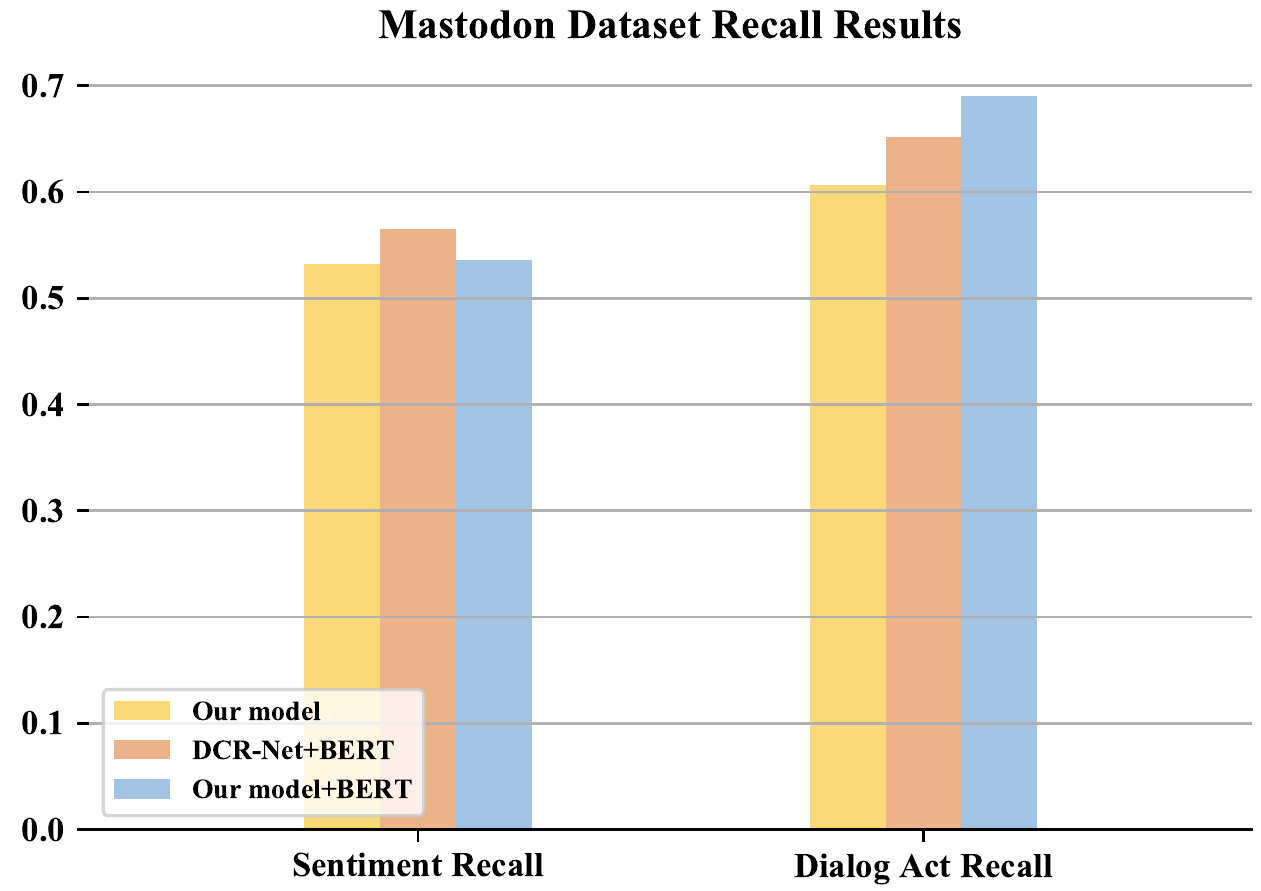}\\
		(b) 
	\end{minipage}
	\begin{minipage}{0.33\linewidth}\centering
		\includegraphics[width=4.8cm]{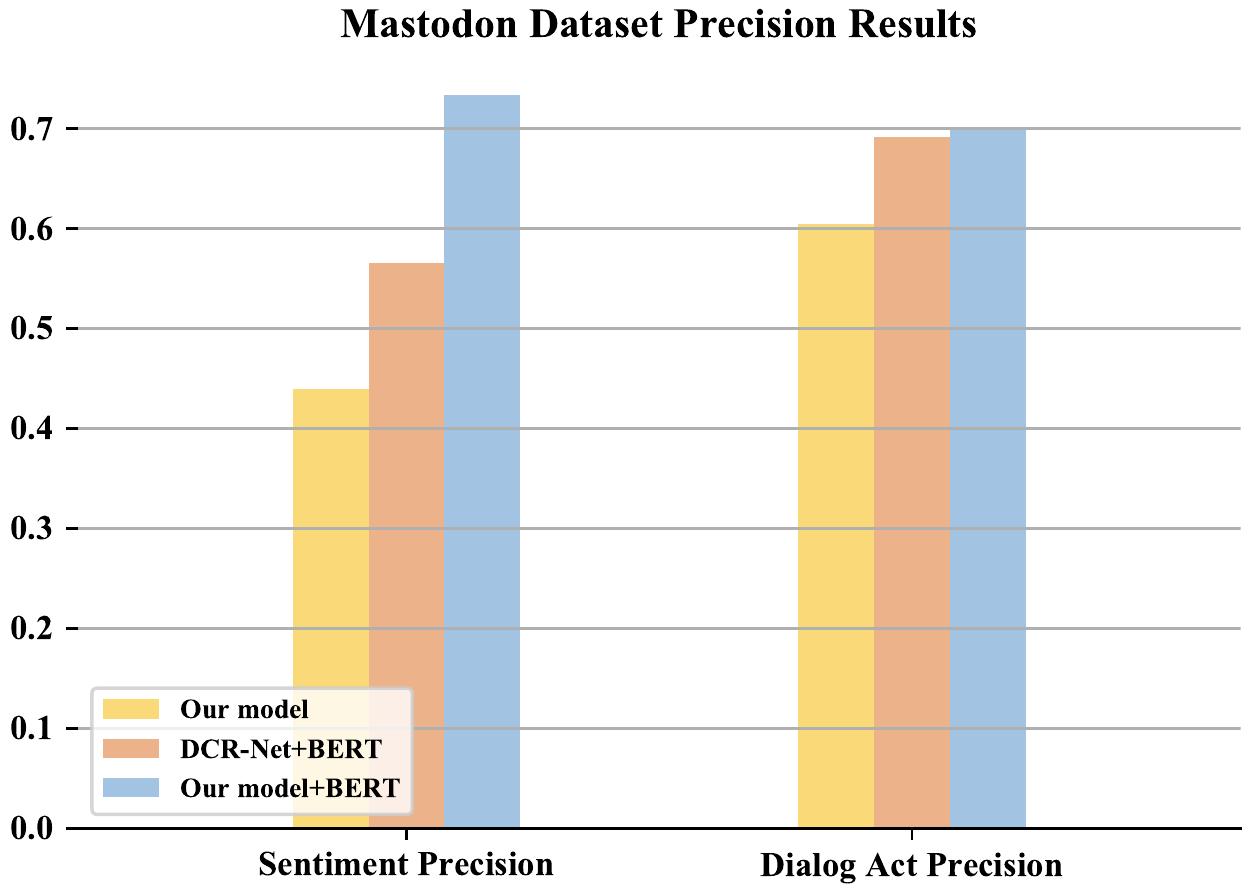}\\
		(c) 
	\end{minipage}
	\vspace{-0.1in}
	\caption{The performance of BERT-based model on Mastodon.}
	\label{fig:bert}
\end{figure*}

\subsection{Overall Results}
Following \citet{kim2018integrated,cerisara-etal-2018-multi,qin2020dcr_net},
we adopt macro-average Precision, Recall and F1 for both sentiment classification and dialog act recognition on Dailydialog dataset and 
we adopt the average of the dialog-act specific F1 scores weighted by the prevalence of each dialog act on Mastodon dataset.

The experimental results are shown in Table~\ref{table:over_all}.
The first block of table represents the separate model for dialog act recognition task while the second block denotes the separate model for sentiment classification task.
The third block of table represents the state-of-the-art joint models for the two task.
From the results, we can observe that: 
\begin{enumerate}
	\item Our framework outperforms the state-of-the-art dialog act recognition and sentiment classification models which trained in separate task in all metrics on two datasets.
	This shows that our proposed graph interaction model has incorporated the mutual interaction information between the two tasks which can be effectively utilized for promoting performance mutually. 
	\item We obtain large improvements compared with the state-of-the-art joint models. On Mastodon dataset, compared with \textit{DCR-Net} model, our framework achieves 3.0\% improvement on F1 score on sentiment classification task and 1.9\% improvement on F1 score on dialog act recognition task.
	On Dailydialog dataset, the same trend has been observed. This demonstrates the effectiveness of simultaneously leveraging contextual information and the mutual interaction information with graph-interaction method, compared with \textit{DCR-Net} which separate considers the two types of information.
\end{enumerate}

\begin{table}[t]
	\small
	\centering
	\begin{adjustbox}{width=0.45\textwidth}
		\begin{tabular}{l|ccc|ccc} 
			\hline 
			\multirow{3}*{\textbf{Model}} & \multicolumn{6}{c}{\textbf{Mastodon}} \\ 
			\cline{2-7} 
			~ & \multicolumn{3}{c|}{SC} & \multicolumn{3}{c}{DAR} \\ 
			\cline{2-7}  
			~ & F1 (\%) & R (\%) & P (\%) & F1 (\%) & R (\%) & P (\%) \\  
			\hline

			RoBERTa+Linear & 55.7 &54.4   &59.7 &61.6  &61.8  &61.4 \\  
			
			Co-GAT+RoBERTa &61.3 &58.8 & 64.3 & 66.1 & 64.8 & 67.5 \\ 
			\hline
			XLNet+Linear &58.7 &60.9 & 56.6 & 62.6 & 61.8 & 63.4 \\ 
			
			Co-GAT+XLNet & 65.9 & 65.8 & 66.1 & 67.5 & 66.0 & 69.2 \\  
			
			\hline  
			
		\end{tabular}
	\end{adjustbox}
	\caption{Results on the pre-trained models.}
	\label{table:over_all}
	
\end{table}
\subsection{Analysis} \label{sec:analysis}
Although achieving good performance, we would like to know the reason for the improvement. In this section, we study our model from several directions. 
We first conduct several ablation experiments to analyze the effect of different components in our framework, including the effect of mutual interaction and contextual information.
Then, we analyze the effect of simultaneous modeling method.
Next, we incorporate and analyze the pre-trained model (BERT, RoBERTa, XLNet) in our framework.

\noindent \textbf{Effectiveness of the Mutual Interaction Information}
In this setting, when constructing the graph architecture for graph interaction, we only consider the \textit{cross-utterances connection} by removing the edges connecting from one task to another task, which can be seen as ignoring the mutual interaction information.
We name it as \textit{without cross-tasks connection} and 
the result is shown in Table~\ref{table:no_relation}.
We can see 2.0\% and 1.3\% drop in terms of F1 scores in sentiment classification while 2.4\% and 1.2\% drop in dialog act recognition on Mastodon and Dailydialog dataset, respectively.
We attribute it to the fact that explicitly modeling the interaction between two tasks with graph-interaction layer can encourage model to effectively utilize the information of one task for another task.

\noindent \textbf{Effectiveness of the Contextual Information}
Similarly, when constructing the graph architecture for graph-interaction, we only consider the \textit{cross-tasks connection} by removing the edges connecting from one node to its contextual node, which can be seen as ignoring the contextual information.
We name it as \textit{without cross-utterances connection} and the result is shown in Table~\ref{table:no_relation}.
The results show a significant drop in
performance, which verifies the effectiveness of contextual information.
The reason is that contextual information help reduce ambiguity, which improves performance. 

\noindent \textbf{Simultaneous Modeling vs. Separate Modeling}
To verify the effectiveness of simultaneously modeling the two sources of information in a unified co-interactive graph interaction mechanism, we remove the co-interactive interaction layer and only use two separate sub GAT to represent the \textit{cross-utterance connection} and \textit{cross-task connection} to model the two tasks separately and adopt the sum operation based on the output of GAT to consider their interaction.
We refer it as \textit{separate modeling} and 
the result is shown in Table~\ref{table:no_relation} and 
the results show a significant drop in
performance.
This indicates that modeling the two sources of information with a co-interactive graph interaction mechanism can better incorporate information simultaneously compared with model the two types of information separately.

In particular, \textit{DCR-Net} can be seen as the SOTA \textit{pipeline} method. To make a more fair comparison with \textit{DCR-Net}, we replace the co-interactive interaction layer with co-attention mechanism in \textit{DCR-Net} and we keep other components unchanged. 
We name it as \textit{co-attention mechanism}.
The results are shown in Table~\ref{table:no_relation} and we can see that our framework outperforms the \textit{co-attention mechanism} by a large margin.
This again demonstrates that simultaneously modeling the contextual information and interaction information by proposed co-interactive graph interaction mechanism is effective than the \textit{pipeline} model to incorporate two types of information in \textit{DCR-Net}.

\noindent \textbf{Effectiveness of Speaker Information}
In this settings, we remove the speaker-aware encoder and only keep the  BiLSTM encoder as the same.
We refer it as \textit{without speaker information} and the result is shown in Table~\ref{table:no_relation}.
From the result, we can see that 1.7\% and 3.4\% drop in terms of F1 scores in sentiment classification while 1.5\% and 0.2\% drop in dialog act recognition on two datasets.
On Dailydialog dataset, we can also observe the same trends that the F1 score drops a lot. 
This demonstrates that properly modeling the speaker information can help model to capture the sentiment and act flow in a dialog, which can enhance their performance.
It is noticeable that even without the speaker-aware encoder, our framework still performs the state-of-the-art \textit{DCR-Net} model , which again demonstrates the effectiveness and robustness of our framework.

\noindent \textbf{Effectiveness of Pre-trained Model}
Finally, following \citet{qin2020dcr_net}, we also explore the pre-trained model, BERT \cite{devlin-etal-2019-bert} in our framework. In this section, we replace the hierarchical speaker-aware encoder by BERT base model \footnote{The BERT is fine-tuned in our experiment.} and keep other components as same with our framework. 
We conduct experiments on Mastodon dataset and the results are shown in Figure~\ref{fig:bert}.
From the results, we can observe: 1) the BERT-based model performs remarkably well and achieves a new state-of-the-art performances. 
This indicates that the performance can be further improved a lot with the pre-trained model and our framework works orthogonally with BERT.
We attribute this to the fact that pre-trained models can provide rich semantic features, which can improve the performance on both two tasks.
2) Our BERT-based model outperforms the baseline (\textit{DCR-Net} + BERT), which again verifies the effectiveness of our proposed co-interactive graph interaction framework.

In addition, to further verify the contribution from our proposed model is still effective over the strong pre-trained model, we perform experiments with Roberta and XLNet.
To further verify that our contribution from Co-GAT does not fully overlap with contextualized word representations (Roberta, XLNet), we have conducted the following experiments on Mastodon dataset:

1)	Roberta/XLNet+Linear. In this setting, we adopt the Roberta/XLNet model as the shared encoder and add two different linear decoders for SC and DAR task. 

2)	Co-GAT + Roberta/XLNet. Here, we replace the hierarchical speaker-aware encoder by Roberta/XLNet model and keep other components as same with our framework. The Roberta/XLNet is fine-tuned in our experiment.

Results are shown in Table 3.
From the results, we find that the integration of Co-GAT and Roberta/XLNet can further improve the performance, demonstrating that contributions from the two are complementary.

\section{Related Work}
\label{RelatedWork}

\subsection{Dialog Act Recognition}
\citet{kalchbrenner2013recurrent} propose a  hierarchical CNN to model the context information for DAR.
\citet{lee2016sequential} propose a model which combine the advantages of CNNs and RNNs and incorporated the previous utterance as context to classify the current for DAR.
\citet{ji2016latent} use a hybrid architecture, combining an RNN language model with a latent variable model.
Furthermore, many work \cite{liu-etal-2017-using-context,kumar2018dialogue,chen2018dialogue} explore different architectures to better incorporate the context information for DAR.
\citet{raheja-tetreault-2019-dialogue} propose the context-aware self-attention mechanism for DAR and achieve the promising performance.
\subsection{Sentiment Classification}
Sentiment classification in dialog system can be seen as the sentence-level sequence classification problem.
One series of works are based on CNN \cite{zhang2015character,conneau-etal-2017-deep,johnson-zhang-2017-deep} to capture the local correlation and position-invariance.
Another series of works adopt RNN based models \cite{tang-etal-2015-document,yang-etal-2016-hierarchical,xu-etal-2016-cached} to capture temporal features for sentiment classification.
Besides, Some works \cite{xiao2016efficient,shi-etal-2016-deep,wang-2018-disconnected} combine the advantages of CNN and RNN.
Recently, \citet{majumder2019dialoguernn} introduce a party state and global state based recurrent model for modeling the emotional dynamics.
\citet{majumder2019dialoguernn} propose a dialogGCN which leverages self and inter-speaker dependency of the interlocutors to model context and achieve the state-of-the-art performance.

\subsection{Joint Model}
Considering the correlation between dialog act recognition and sentiment classification,
many joint models are proposed to
consider the interaction between the two tasks.
\citet{cerisara-etal-2018-multi} explore the multi-task framework to model the correlation between the two tasks.
\citet{kim2018integrated} propose an integrated 
neural network for identifying dialog act, predicators, and sentiments of dialogue utterances.
Compared with their work, our framework 
simultaneously considers the contextual information and mutual interaction information into a unified graph interaction architecture. In contrast, their models only consider on type of information (contextual information or mutual interaction information).
More recently, \citet{qin2020dcr_net} propose a DCR-Net which adopts a relation layer to model the relationship and achieve the state-of-the-art performance.
This model can be regarded as the \textit{pipeline} method to model the contextual and mutual interaction information, which ignores the contextual information when performing interaction between the two tasks.
In contrast, we propose a co-interactive graph attention network where \textit{cross-utterances connection} and \textit{cross-tasks connection} are constructed and iteratively updated with each other to  simultaneously model the contextual information and the mutual interaction information into a unified graph structure. 
To the best of our knowledge, we are the first to simultaneously model the mutual information and contextual information in a unified graph interaction architecture

\section{Conclusion}
\label{Conclusion}
In this paper, we propose a co-interactive graph framework where a \textit{cross-utterances connection} and a \textit{cross-tasks connection} are constructed and iteratively updated with each other, achieving to simultaneously model the contextual information and mutual interaction information in a unified architecture.
Experiments on two datasets show the effectiveness of the proposed models and our model achieves state-of-the-art performance. 
 In addition, we analyze the effect of incorporating strong pre-trained model in our joint model and find that our framework is also beneficial when combined with pre-trained models (BERT, Roberta, XLNet). 
 
 \section{Acknowledgments}
 We thank the anonymous reviewers for their helpful comments and suggestions. This work was supported by the National Key R\&D Program of China via grant 2020AAA0106501 and the National Natural Science Foundation of China (NSFC) via grant 61976072 and 61772153.



\bibliography{aaai21}

\begin{thebibliography}{36}
\providecommand{\natexlab}[1]{#1}
\providecommand{\url}[1]{\texttt{#1}}
\providecommand{\urlprefix}{URL }
\expandafter\ifx\csname urlstyle\endcsname\relax
  \providecommand{\doi}[1]{doi:\discretionary{}{}{}#1}\else
  \providecommand{\doi}{doi:\discretionary{}{}{}\begingroup
  \urlstyle{rm}\Url}\fi

\bibitem[{Cerisara et~al.(2018)Cerisara, Jafaritazehjani, Oluokun, and
  Le}]{cerisara-etal-2018-multi}
Cerisara, C.; Jafaritazehjani, S.; Oluokun, A.; and Le, H.~T. 2018.
\newblock Multi-task dialog act and sentiment recognition on Mastodon.
\newblock In \emph{Proceedings of the 27th International Conference on
  Computational Linguistics}, 745--754. Santa Fe, New Mexico, USA: Association
  for Computational Linguistics.
\newblock \urlprefix\url{https://www.aclweb.org/anthology/C18-1063}.

\bibitem[{Chai and Wan(2020)}]{chai-wan-2020-learning}
Chai, Z.; and Wan, X. 2020.
\newblock Learning to Ask More: Semi-Autoregressive Sequential Question
  Generation under Dual-Graph Interaction.
\newblock In \emph{Proceedings of the 58th Annual Meeting of the Association
  for Computational Linguistics}, 225--237. Online: Association for
  Computational Linguistics.
\newblock \doi{10.18653/v1/2020.acl-main.21}.
\newblock \urlprefix\url{https://www.aclweb.org/anthology/2020.acl-main.21}.

\bibitem[{Chen et~al.(2018)Chen, Yang, Zhao, Cai, and He}]{chen2018dialogue}
Chen, Z.; Yang, R.; Zhao, Z.; Cai, D.; and He, X. 2018.
\newblock Dialogue act recognition via crf-attentive structured network.
\newblock In \emph{Proc. of SIGIR}.

\bibitem[{Conneau et~al.(2017)Conneau, Schwenk, Barrault, and
  Lecun}]{conneau-etal-2017-deep}
Conneau, A.; Schwenk, H.; Barrault, L.; and Lecun, Y. 2017.
\newblock Very Deep Convolutional Networks for Text Classification.
\newblock In \emph{Proceedings of the 15th Conference of the {E}uropean Chapter
  of the Association for Computational Linguistics: Volume 1, Long Papers},
  1107--1116. Valencia, Spain: Association for Computational Linguistics.
\newblock \urlprefix\url{https://www.aclweb.org/anthology/E17-1104}.

\bibitem[{Devlin et~al.(2019)Devlin, Chang, Lee, and
  Toutanova}]{devlin-etal-2019-bert}
Devlin, J.; Chang, M.-W.; Lee, K.; and Toutanova, K. 2019.
\newblock {BERT}: Pre-training of Deep Bidirectional Transformers for Language
  Understanding.
\newblock In \emph{Proc. of NAACL}.

\bibitem[{Ghosal et~al.(2019)Ghosal, Majumder, Poria, Chhaya, and
  Gelbukh}]{ghosal-etal-2019-dialoguegcn}
Ghosal, D.; Majumder, N.; Poria, S.; Chhaya, N.; and Gelbukh, A. 2019.
\newblock {D}ialogue{GCN}: A Graph Convolutional Neural Network for Emotion
  Recognition in Conversation.
\newblock In \emph{Proceedings of the 2019 Conference on Empirical Methods in
  Natural Language Processing and the 9th International Joint Conference on
  Natural Language Processing (EMNLP-IJCNLP)}, 154--164. Hong Kong, China:
  Association for Computational Linguistics.
\newblock \doi{10.18653/v1/D19-1015}.
\newblock \urlprefix\url{https://www.aclweb.org/anthology/D19-1015}.

\bibitem[{Hochreiter and Schmidhuber(1997)}]{hochreiter1997long}
Hochreiter, S.; and Schmidhuber, J. 1997.
\newblock Long short-term memory.
\newblock \emph{Neural computation} .

\bibitem[{Ji, Haffari, and Eisenstein(2016)}]{ji2016latent}
Ji, Y.; Haffari, G.; and Eisenstein, J. 2016.
\newblock A latent variable recurrent neural network for discourse relation
  language models.
\newblock \emph{arXiv preprint arXiv:1603.01913} .

\bibitem[{Johnson and Zhang(2017)}]{johnson-zhang-2017-deep}
Johnson, R.; and Zhang, T. 2017.
\newblock Deep Pyramid Convolutional Neural Networks for Text Categorization.
\newblock In \emph{Proceedings of the 55th Annual Meeting of the Association
  for Computational Linguistics (Volume 1: Long Papers)}, 562--570. Vancouver,
  Canada: Association for Computational Linguistics.
\newblock \doi{10.18653/v1/P17-1052}.
\newblock \urlprefix\url{https://www.aclweb.org/anthology/P17-1052}.

\bibitem[{Kalchbrenner and Blunsom(2013)}]{kalchbrenner2013recurrent}
Kalchbrenner, N.; and Blunsom, P. 2013.
\newblock Recurrent convolutional neural networks for discourse
  compositionality.
\newblock \emph{arXiv preprint arXiv:1306.3584} .

\bibitem[{Kim and Kim(2018)}]{kim2018integrated}
Kim, M.; and Kim, H. 2018.
\newblock Integrated neural network model for identifying speech acts,
  predicators, and sentiments of dialogue utterances.
\newblock \emph{Pattern Recognition Letters} .

\bibitem[{Kingma and Ba(2014)}]{kingma-ba:2014:ICLR}
Kingma, D.~P.; and Ba, J. 2014.
\newblock Adam: A method for stochastic optimization.
\newblock \emph{arXiv preprint arXiv:1412.6980} .

\bibitem[{Kumar et~al.(2018)Kumar, Agarwal, Dasgupta, and
  Joshi}]{kumar2018dialogue}
Kumar, H.; Agarwal, A.; Dasgupta, R.; and Joshi, S. 2018.
\newblock Dialogue act sequence labeling using hierarchical encoder with crf.
\newblock In \emph{Proc. of AAAI}.

\bibitem[{Lee and Dernoncourt(2016)}]{lee2016sequential}
Lee, J.~Y.; and Dernoncourt, F. 2016.
\newblock Sequential short-text classification with recurrent and convolutional
  neural networks.
\newblock \emph{arXiv preprint arXiv:1603.03827} .

\bibitem[{Li et~al.(2017)Li, Su, Shen, Li, Cao, and
  Niu}]{li-etal-2017-dailydialog}
Li, Y.; Su, H.; Shen, X.; Li, W.; Cao, Z.; and Niu, S. 2017.
\newblock {D}aily{D}ialog: A Manually Labelled Multi-turn Dialogue Dataset.
\newblock In \emph{Proceedings of the Eighth International Joint Conference on
  Natural Language Processing (Volume 1: Long Papers)}, 986--995. Taipei,
  Taiwan: Asian Federation of Natural Language Processing.
\newblock \urlprefix\url{https://www.aclweb.org/anthology/I17-1099}.

\bibitem[{Lin, Xu, and Zhang(2020)}]{lin2020discovering}
Lin, T.-E.; Xu, H.; and Zhang, H. 2020.
\newblock Discovering New Intents via Constrained Deep Adaptive Clustering with
  Cluster Refinement.
\newblock In \emph{Thirty-Fourth AAAI Conference on Artificial Intelligence}.

\bibitem[{Liu et~al.(2017)Liu, Han, Tan, and Lei}]{liu-etal-2017-using-context}
Liu, Y.; Han, K.; Tan, Z.; and Lei, Y. 2017.
\newblock Using Context Information for Dialog Act Classification in {DNN}
  Framework.
\newblock In \emph{Proceedings of the 2017 Conference on Empirical Methods in
  Natural Language Processing}, 2170--2178. Copenhagen, Denmark: Association
  for Computational Linguistics.
\newblock \doi{10.18653/v1/D17-1231}.
\newblock \urlprefix\url{https://www.aclweb.org/anthology/D17-1231}.

\bibitem[{Liu et~al.(2019)Liu, Ott, Goyal, Du, Joshi, Chen, Levy, Lewis,
  Zettlemoyer, and Stoyanov}]{liu2019roberta}
Liu, Y.; Ott, M.; Goyal, N.; Du, J.; Joshi, M.; Chen, D.; Levy, O.; Lewis, M.;
  Zettlemoyer, L.; and Stoyanov, V. 2019.
\newblock Roberta: A robustly optimized bert pretraining approach.
\newblock \emph{arXiv preprint arXiv:1907.11692} .

\bibitem[{Lu and Li(2020)}]{lu-li-2020-gcan}
Lu, Y.-J.; and Li, C.-T. 2020.
\newblock {GCAN}: Graph-aware Co-Attention Networks for Explainable Fake News
  Detection on Social Media.
\newblock In \emph{Proceedings of the 58th Annual Meeting of the Association
  for Computational Linguistics}, 505--514. Online: Association for
  Computational Linguistics.
\newblock \doi{10.18653/v1/2020.acl-main.48}.
\newblock \urlprefix\url{https://www.aclweb.org/anthology/2020.acl-main.48}.

\bibitem[{Majumder et~al.(2019)Majumder, Poria, Hazarika, Mihalcea, Gelbukh,
  and Cambria}]{majumder2019dialoguernn}
Majumder, N.; Poria, S.; Hazarika, D.; Mihalcea, R.; Gelbukh, A.; and Cambria,
  E. 2019.
\newblock Dialoguernn: An attentive rnn for emotion detection in conversations.
\newblock In \emph{Proceedings of the AAAI Conference on Artificial
  Intelligence}.

\bibitem[{Qin et~al.(2020{\natexlab{a}})Qin, Che, Li, Ni, and
  Liu}]{qin2020dcr_net}
Qin, L.; Che, W.; Li, Y.; Ni, M.; and Liu, T. 2020{\natexlab{a}}.
\newblock DCR-Net: A Deep Co-Interactive Relation Network for Joint Dialog Act
  Recognition and Sentiment Classification.
\newblock In \emph{Proceedings of the AAAI Conference on Artificial
  Intelligence}.

\bibitem[{Qin et~al.(2019)Qin, Che, Li, Wen, and Liu}]{qin-etal-2019-stack}
Qin, L.; Che, W.; Li, Y.; Wen, H.; and Liu, T. 2019.
\newblock A Stack-Propagation Framework with Token-Level Intent Detection for
  Spoken Language Understanding.
\newblock In \emph{Proc. of EMNLP}.

\bibitem[{Qin et~al.(2020{\natexlab{b}})Qin, Ni, Zhang, Che, Li, and
  Liu}]{qin2020multi}
Qin, L.; Ni, M.; Zhang, Y.; Che, W.; Li, Y.; and Liu, T. 2020{\natexlab{b}}.
\newblock Multi-Domain Spoken Language Understanding Using Domain-and
  Task-Aware Parameterization.
\newblock \emph{arXiv preprint arXiv:2004.14871} .

\bibitem[{Qin et~al.(2020{\natexlab{c}})Qin, Xu, Che, and
  Liu}]{qin-etal-2020-agif}
Qin, L.; Xu, X.; Che, W.; and Liu, T. 2020{\natexlab{c}}.
\newblock {AGIF}: An Adaptive Graph-Interactive Framework for Joint Multiple
  Intent Detection and Slot Filling.
\newblock In \emph{Findings of the Association for Computational Linguistics:
  EMNLP 2020}, 1807--1816. Online: Association for Computational Linguistics.
\newblock \doi{10.18653/v1/2020.findings-emnlp.163}.
\newblock
  \urlprefix\url{https://www.aclweb.org/anthology/2020.findings-emnlp.163}.

\bibitem[{Raheja and Tetreault(2019)}]{raheja-tetreault-2019-dialogue}
Raheja, V.; and Tetreault, J. 2019.
\newblock {D}ialogue {A}ct {C}lassification with {C}ontext-{A}ware
  {S}elf-{A}ttention.
\newblock In \emph{Proc. of NAACL}.

\bibitem[{Scarselli et~al.(2009)Scarselli, Gori, Tsoi, Hagenbuchner, and
  Monfardini}]{scarselli2009graph}
Scarselli, F.; Gori, M.; Tsoi, A.~C.; Hagenbuchner, M.; and Monfardini, G.
  2009.
\newblock The graph neural network model.
\newblock \emph{IEEE Transactions on Neural Networks} 20(1): 61--80.

\bibitem[{Shi et~al.(2016)Shi, Yao, Tian, and Jiang}]{shi-etal-2016-deep}
Shi, Y.; Yao, K.; Tian, L.; and Jiang, D. 2016.
\newblock Deep {LSTM} based Feature Mapping for Query Classification.
\newblock In \emph{Proc. of NAACL}.

\bibitem[{Tang, Qin, and Liu(2015)}]{tang-etal-2015-document}
Tang, D.; Qin, B.; and Liu, T. 2015.
\newblock Document Modeling with Gated Recurrent Neural Network for Sentiment
  Classification.
\newblock In \emph{Proc. of ACL}.

\bibitem[{Vaswani et~al.(2017)Vaswani, Shazeer, Parmar, Uszkoreit, Jones,
  Gomez, Kaiser, and Polosukhin}]{NIPS2017_7181}
Vaswani, A.; Shazeer, N.; Parmar, N.; Uszkoreit, J.; Jones, L.; Gomez, A.~N.;
  Kaiser, L.~u.; and Polosukhin, I. 2017.
\newblock Attention is All you Need.
\newblock In \emph{Proc. of NIPS}. Curran Associates, Inc.

\bibitem[{Veli{\v{c}}kovi{\'c} et~al.(2017)Veli{\v{c}}kovi{\'c}, Cucurull,
  Casanova, Romero, Lio, and Bengio}]{velivckovic2017graph}
Veli{\v{c}}kovi{\'c}, P.; Cucurull, G.; Casanova, A.; Romero, A.; Lio, P.; and
  Bengio, Y. 2017.
\newblock Graph attention networks.
\newblock \emph{arXiv preprint arXiv:1710.10903} .

\bibitem[{Wang(2018)}]{wang-2018-disconnected}
Wang, B. 2018.
\newblock Disconnected Recurrent Neural Networks for Text Categorization.
\newblock In \emph{Proc. of ACL}.

\bibitem[{Xiao and Cho(2016)}]{xiao2016efficient}
Xiao, Y.; and Cho, K. 2016.
\newblock Efficient character-level document classification by combining
  convolution and recurrent layers.
\newblock \emph{arXiv preprint arXiv:1602.00367} .

\bibitem[{Xu et~al.(2016)Xu, Chen, Qiu, and Huang}]{xu-etal-2016-cached}
Xu, J.; Chen, D.; Qiu, X.; and Huang, X. 2016.
\newblock Cached Long Short-Term Memory Neural Networks for Document-Level
  Sentiment Classification.
\newblock In \emph{Proc. of EMNLP}.

\bibitem[{Yang et~al.(2019)Yang, Dai, Yang, Carbonell, Salakhutdinov, and
  Le}]{yang2019xlnet}
Yang, Z.; Dai, Z.; Yang, Y.; Carbonell, J.; Salakhutdinov, R.~R.; and Le, Q.~V.
  2019.
\newblock Xlnet: Generalized autoregressive pretraining for language
  understanding.
\newblock In \emph{Advances in neural information processing systems},
  5753--5763.

\bibitem[{Yang et~al.(2016)Yang, Yang, Dyer, He, Smola, and
  Hovy}]{yang-etal-2016-hierarchical}
Yang, Z.; Yang, D.; Dyer, C.; He, X.; Smola, A.; and Hovy, E. 2016.
\newblock Hierarchical Attention Networks for Document Classification.
\newblock In \emph{Proc. of NAACL}.

\bibitem[{Zhang, Zhao, and LeCun(2015)}]{zhang2015character}
Zhang, X.; Zhao, J.; and LeCun, Y. 2015.
\newblock Character-level convolutional networks for text classification.
\newblock In \emph{Proc. of NIPS}, 649--657.

\end{thebibliography}
\end{document}